\documentclass[10pt,twocolumn,letterpaper]{article}

\usepackage{cvpr,times}

\usepackage{graphicx,amsmath,amssymb,subcaption}

\usepackage{url}
\usepackage{xcolor}
\definecolor{newcolor}{rgb}{0.8,0.349,0.1}

\usepackage{hyperref,cleveref}
\hypersetup{colorlinks=true,linkcolor=blue,urlcolor=blue,citecolor=blue}
\hypersetup{pagebackref=true,breaklinks=true,letterpaper=true,bookmarks=false}

\cvprfinalcopy

\def\cvprPaperID{52}

\begin{document}

\title{A Comprehensive Analysis of Deep Learning Based Representation \\ for Face Recognition}

\author{Mostafa Mehdipour Ghazi\\
Faculty of Engineering and Natural Sciences\\
Sabanci University, Istanbul, Turkey\\
{\tt\small mehdipour@sabanciuniv.edu}
\and
Haz{\i}m Kemal Ekenel\\
Department of Computer Engineering\\
Istanbul Technical University, Istanbul, Turkey\\
{\tt\small ekenel@itu.edu.tr}
}

\maketitle

\begin{abstract}
Deep learning based approaches have been dominating the face recognition field due to the significant performance improvement they have provided on the challenging wild datasets. These approaches have been extensively tested on such unconstrained datasets, on the Labeled Faces in the Wild and YouTube Faces, to name a few. However, their capability to handle individual appearance variations caused by factors such as head pose, illumination, occlusion, and misalignment has not been thoroughly assessed till now. In this paper, we present a comprehensive study to evaluate the performance of deep learning based face representation under several conditions including the varying head pose angles, upper and lower face occlusion, changing illumination of different strengths, and misalignment due to erroneous facial feature localization. Two successful and publicly available deep learning models, namely VGG-Face and Lightened CNN have been utilized to extract face representations. The obtained results show that although deep learning provides a powerful representation for face recognition, it can still benefit from preprocessing, for example, for pose and illumination normalization in order to achieve better performance under various conditions. Particularly, if these variations are not included in the dataset used to train the deep learning model, the role of preprocessing becomes more crucial. Experimental results also show that deep learning based representation is robust to misalignment and can tolerate facial feature localization errors up to 10\% of the interocular distance.
\end{abstract}

\section{Introduction}
Human face recognition is a challenging problem in computer vision with several biometrics applications. This problem essentially faces difficulties due to variations in facial appearance caused by factors such as illumination, expression, and partial occlusion from accessories including glasses, scarves, hats, and the like.

In recent years, deep learning based approaches have been increasingly applied for face recognition with promising results \cite{deepface2014, deepid2014, facenet2015, vggface2015, lightened2015}. These methods take raw data as their network input and convolve filters in multiple levels to automatically discover low-level and high-level representations from labeled or unlabeled data for detecting, distinguishing, and/or classifying their underlying patterns \cite{hinton2006, kavukcuoglu2010, alexnet2012, googlenet2015, lecun2015}. However, optimizing millions of parameters to learn the multi-stage weights from scratch in deep learning architectures requires millions of training samples and an access to powerful computational resources such as Graphical Processing Units (GPUs). Consequently, the method of transfer learning \cite{torrey2009, pan2010} is efficiently utilized to apply previously learned knowledge of a relevant visual recognition problem to the new, desired task domain.

Transfer learning can be applied in two different ways with respect to the size and similarity between the pre-training dataset and the new database. The first approach is fine-tuning the pre-trained network weights using the new dataset via backpropagation. This method is only suggested for large enough datasets since fine-tuning the pre-trained networks with few training samples can lead to overfitting \cite{yosinski2014}. The second approach is the direct utilization of learned weights in the desired problem to extract and later classify features. This scheme is especially efficient when the new dataset is small and/or a few number of classes exists. Depending on the task similarity between the two datasets, one can decide whether to use lower layers' weights--as generic low-level feature extractors--or higher layers' weights--as task specific motif extractors \cite{lecun2015}.

In this paper, the higher layer portion of learned weights from two deep convolutional neural networks (CNNs) of VGG-Face \cite{vggface2015} and Lightened CNN \cite{lightened2015}, pre-trained on very large face recognition collections, have been employed to extract face representation. These two models are selected since they have been found to be successful for face recognition in the wild while being publicly available. The former network includes a very deep architecture and the latter is a computationally efficient CNN. Robustness of these deep face representations against variations of different factors including illumination, occlusion, pose, and misalignment has been thoroughly assessed using five popular face datasets, namely the AR \cite{ar1998}, CMU PIE \cite{pie2002}, Extended Yale dataset \cite{extyale2001}, Color FERET \cite{feret1998}, and FRGC \cite{frgc2005}.

The main contributions and outcomes of this work can be summarized as follows: (i) A comprehensive evaluation of deep learning based representation under various conditions including pose, illumination, occlusion, and misalignment has been conducted. In fact, all the proposed deep learning based face recognition methods such as DeepFace \cite{deepface2014}, DeepID \cite{deepid2014}, FaceNet \cite{facenet2015}, and VGG-Face \cite{vggface2015} have been trained and evaluated on very large wild face recognition datasets, i.e. Labeled Faces in the Wild (LFW) \cite{lfw2007}, YouTube Faces (YTF) \cite{ytf2011}, and MegaFace \cite{megaface2015}. However, their representation capabilities to handle individual appearance variations have not been assessed yet. (ii) We have shown that although deep learning provides a powerful representation for face recognition, it is not able to achieve state-of-the-art results against pose, illumination, and occlusion. To enable deep learning models achieve better results, either these variations should be taken into account during training or preprocessing methods for pose and illumination normalization should be employed along with pre-trained models. (iii) We have found that deep learning based face representation is robust to misalignment and able to tolerate facial feature localization errors up to 10\% of the interocular distance. (iv) The VGG-Face model \cite{vggface2015} is shown to be more transferable compared to the Lightened CNN model \cite{lightened2015}. Overall, we believe that deep learning based face recognition requires further research to address the problem of face recognition under mismatched conditions, especially when there is a limited amount of data available for the task at hand.

The rest of the paper is organized as follows. \Cref{related_sec} covers a review of existing deep learning methods for face recognition. \Cref{method_sec} describes the details of two deep CNN models for face recognition and presents the extraction and assessment approach for face representation based on these models. \Cref{result_sec} explains the utilized datasets and presents the designed experiments and their results. Finally, \Cref{conc_sec} concludes the paper with the summary and discussion of the conducted experiments and implications of the obtained results.

\section{Related Work} \label{related_sec}
Before the emergence of deep learning algorithms, the majority of traditional face recognition methods used to first locally extract hand-crafted shallow features from facial images using Local Binary Patterns (LBP), Scale Invariant Feature Transform (SIFT), and Histogram of Oriented Gradients (HOG), and later train features and classify identities by Support Vector Machines (SVMs) or Nearest Neighbors (NNs) \cite{ahonen2004, geng2009, sivic2009, deniz2011}. However, with the availability of the state-of-the-art computational resources and with a surge in access to very large datasets, deep learning architectures have been developed and shown immensely impressive results for different visual recognition tasks including face recognition \cite{deepface2014, deepid2014, facenet2015, vggface2015, lightened2015}.

DeepFace \cite{deepface2014} is one of these outstanding networks that contains a nine-layer deep CNN model with two convolutional layers and more than 120 million parameters trained on four million facial images from over 4,000 identities. This method, through alignment of images based on a 3D model and use of an ensemble of CNNs, could achieve accuracies of 97.35\% and 91.4\% on the LFW and YTF datasets, respectively. Deep hidden IDentity features (DeepID) \cite{deepid2014} is another successful deep learning method proposed for face recognition and verification with a nine-layer network and four convolutional layers. This scheme first learns weights through face identification and extracts features using the last hidden layer outputs, and later generalizes them to face verification. DeepID aligns faces by similarity transformation based on two eye centers and two mouth corners. This network was trained on the Celebrity Faces dataset (CelebFaces) \cite{celebfaces2013} and achieved an accuracy of 97.45\% on the LFW dataset.

FaceNet \cite{facenet2015} is a deep CNN based on GoogLeNet \cite{googlenet2015} and the network proposed in \cite{zeiler2014} and trained on a face dataset with 100 to 200 million images of around eight million identities. This algorithm uses triplets of roughly aligned faces obtained from an online triplet mining approach and directly learns to map face images to a compact Euclidean space to measure face similarity. FaceNet has been evaluated on the LFW and YTF datasets and has achieved accuracies of 99.63\% and 95.12\%, respectively.

\section{Methods} \label{method_sec}
In this section, we present and describe two successful CNN architectures for face recognition and discuss face representation based on these models.

\subsection{VGG-Face Network}
VGG-Face \cite{vggface2015} is a deep convolutional network proposed for face recognition using the VGGNet architecture \cite{vggnet2014}. It is trained on 2.6 million facial images of 2,622 identities collected from the web. The network involves 16 convolutional layers, five max-pooling layers, three fully-connected layers, and a final linear layer with Softmax activation. VGG-Face takes color image patches of size $224 \times 224$ pixels as the input and utilizes dropout regularization \cite{dropout2014} in the fully-connected layers. Moreover, it applies ReLU activation to all of its convolutional layers. Spanning 144 million parameters clearly reveals that the VGG network is a computationally expensive architecture. This method has been evaluated on the LFW dataset and achieved an accuracy of 98.95\%.

\subsection{Lightened CNN}
This framework is a CNN with a low computational complexity proposed for face recognition \cite{lightened2015}. It uses an activation function called Max-Feature-Map (MFM) to extract more abstract representations in comparison with the ReLU. Lightened CNN is introduced in two different models. The first network, (A), inspired by the AlexNet model \cite{alexnet2012}, contains 3,961K parameters with four convolution layers using the MFM activation functions, four max-pooling layers, two fully-connected layers, and a linear layer with Softmax activation in the output. The second network, (B), is inspired by the Network in Network model \cite{nin2013} and involves 3,244K parameters with five convolution layers using the MFM activation functions, four convolutional layers for dimensionality reduction, five max-pooling layers, two fully-connected layers, and a linear layer with Softmax activation in the output.

The Lightened CNN models take grayscale facial patch images of size $128 \times 128$ pixels as the network inputs. These models are trained on 493,456 facial images of 10,575 identities in the CASIA WebFace dataset \cite{casia2014}. Both Lightened CNN models have been evaluated on the LFW dataset and achieved accuracies of 98.13\% and 97.77\%, respectively.

\subsection{Face Representation with CNN Models}
The implemented and pre-trained models of VGG-Face and Lightened CNN are used in the Caffe deep learning framework \cite{caffe2014}. To systematically evaluate robustness of the aforementioned deep CNN models under different appearance variations, all the layer weights of each network until the first fully-connected layer--before the last dropout layer and fully-connected layer with Softmax activation--are used for feature extraction. These layers are indicated as FC6 and FC1 in the VGG-Face and Lightened CNN models, respectively. To analyze the effects of different fully-connected layers, we also deploy the FC7 layer of the VGG-Face network. The VGG-Face model provides a 4096-dimensional, high-level representation extracted from a color image patch of size $224 \times 224$ pixels, whereas the Lightened CNN models provides a 512-dimensional, high-level feature vector obtained from a grayscale image patch of size $128 \times 128$ pixels. Extracted features are then classified using the method of nearest neighbors with cosine distance metric. Although we tested other metrics such as Euclidean distance and cross-correlation as well, the cosine distance almost always achieved the best results.

Preprocessing steps including alignment and/or illumination normalization and contrast enhancement are applied when needed. The face alignment is done with respect to the eye centers while illumination normalization and contrast enhancement are performed using the proposed methods in \cite{tan2010}.

In the conducted experiments, neutral images from the datasets--including face images captured from frontal pose under controlled illumination with no face occlusion--are used for gallery images while probe images contain several appearance variations due to head pose, illumination changes, facial occlusion, and misalignment.

\section{Experiments and Results} \label{result_sec}
In this section, we provide the details of utilized datasets and experimental setups. Furthermore, we present the scenarios used for evaluation of deep CNN-based face representation and discuss the obtained results.

\subsection{The AR Face Database -- Face Occlusion}
The AR face database \cite{ar1998} contains 4,000 color, frontal face images of size $768 \times 576$ pixels with different facial expressions, illuminations, and occlusions from 126 subjects. Each subject had participated in two sessions separated by two weeks and with no restrictions on headwear, make-up, hairstyle, accessories, etc. Since the aim of this experiment is to benchmark the robustness of deep CNN-based features against occlusion, one image per subject with the neutral expression from the first session is used for training. Subsequently, two images per subject per session are used for testing, one while wearing a pair of sunglasses to test the impact of upper face occlusion and one while wearing a scarf to test the effect of lower face occlusion. In total, these samples could be completely acquired from 110 subjects.

\begin{figure}[t]
\centering
\vspace{0.00 in}
\includegraphics[width=0.91\linewidth]{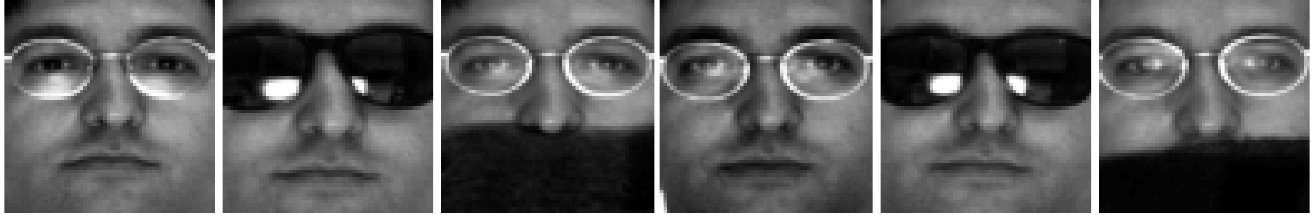}
\caption{Samples from the AR database with different occlusion conditions. The first three images from left are associated with Session 1 and the next three are obtained from Session 2 with repeating conditions of neutral, wearing a pair of sunglasses, and wearing a scarf.}
\label{fig1}
\end{figure}

Each selected image is later aligned, cropped into a square facial patch, and scaled to either $224 \times 224$ or $128 \times 128$ pixels. Finally, the mean image obtained from the training set of VGG-Face is subtracted from each image to ascertain the implementation of the same image transforms applied on pre-tained models. \Cref{fig1} shows images associated with one subject from the AR database used for the experiment. Four experiments are conducted on the AR dataset. The first two experiments involved training and testing within the first session while the rest are trained with samples from the first session and tested on images from the second session. \Cref{table1} summarizes the results of our experiments on occlusion variations using the AR face database.

\begin{table}[t]
\centering
\vspace{0.00 in}
\small
\caption{Classification results (\%) for the AR database using deep features against different occlusion conditions}
\label{table1}
\renewcommand{\arraystretch}{1.2}
\begin{tabular}{cccc} \hline
\textbf{Testing Set} & \multicolumn{2}{c}{\textbf{VGG-Face}} & \textbf{Lightened CNN} \\
 & \textbf{FC6} & \textbf{FC7} & \\ \hline \hline
\text{Sunglasses Session 1} & 33.64 & \textbf{35.45} & 5.45 (A) \\
\text{Scarf Session 1} & 86.36 & \textbf{89.09} & 12.73 (A) \\
\text{Sunglasses Session 2} & \textbf{29.09} & 28.18 & 7.27 (B) \\
\text{Scarf Session 2} & \textbf{85.45} & 83.64 & 10.00 (A) \\ \hline
\end{tabular}
\end{table}

As it can be observed from \Cref{table1}, deep face representation has difficulty to handle upper face occlusion due to wearing sunglasses. Compared to the state-of-the-art occlusion-robust face recognition algorithms \cite{wright2008, ekenel2009, ou2014}, the obtained results with deep representation are rather low. These results indicate that, unless specifically trained on a large amount of data with occlusion, deep CNN-based representation may not function well when a facial occlusion exists. In the same experiments, the VGG-Face model is also found to be more robust against facial occlusion compared to the Lightened CNN models. In this table, only the results of the best performing Lightened CNN models are presented.

\subsection{CMU PIE Database -- Illumination Variations}
The CMU PIE face database \cite{pie2002} contains 41,368 color, facial images of size $640 \times 480$ pixels photographed from 68 subjects under 13 different head poses, 43 different illumination conditions, and four different expressions. Since the goal of the experiment is to evaluate the effects of illumination variations on the performance of deep CNN-based features, frontal images from the illumination subset of the CMU PIE dataset are chosen for further analyses. This subset contains 21 images per subject taken under varying illumination conditions. One frontally illuminated facial image per subject is used for training and the remaining 20 face images containing varying illumination are used for testing.

\begin{figure}[t]
\centering
\vspace{0.00 in}
\includegraphics[width=0.91\linewidth]{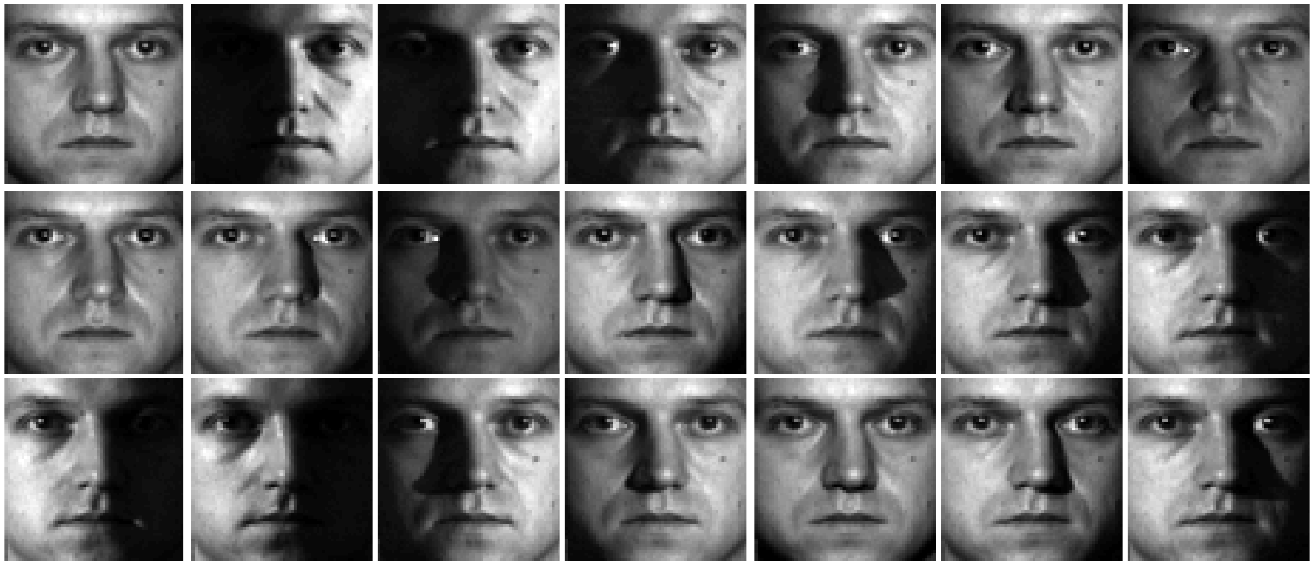}
\caption{Samples from the CMU PIE database with different illumination conditions. The first image in the upper left is the frontal face picture used for training and the rest are assigned for testing.}
\label{fig2}
\end{figure}

All collected images are aligned, cropped into a square facial patch, and finally scaled to either $224 \times 224$ or $128 \times 128$ pixels. The VGG-Face mean image is then subtracted from each image. \Cref{fig2} depicts, as an example, the utilized samples for one subject from the CMU PIE database. Results of the experiments on illumination variations are presented in \Cref{table2}.

As can be seen, the obtained deep representation using the VGG-Face is robust against illumination variations. However, the obtained accuracies are slightly lower compared to the results achieved by the state-of-the-art illumination-robust face recognition approaches \cite{lee2005, zhou2007, tan2010}. These results indicate that the performance of deep face representations needs to be further improved using illumination-based preprocessing methods \cite{tan2010}.

\begin{table}[b]
\centering
\vspace{-0.02 in}
\small
\caption{Classification results (\%) for the CMU PIE database using deep facial features against different illumination conditions}
\label{table2}
\renewcommand{\arraystretch}{1.2}
\begin{tabular}{cccc} \hline
 & \multicolumn{2}{c}{\textbf{VGG-Face}} & \textbf{Lightened CNN} \\
 & \textbf{FC6} & \textbf{FC7} & \\ \hline \hline
\textbf{Accuracy} & \textbf{93.16} & 92.87 & 20.51 (A) \\ \hline
\end{tabular}
\end{table}

\subsection{Extended Yale Dataset -- Illumination Changes}
The extended Yale face dataset B \cite{extyale2001} contains 16,128 images captured from 38 subjects under nine poses and 64 illumination variations. These 64 samples are divided into five subsets according to the angle between the light source direction and camera's optical axis; subset 1 contains seven images with the lighting angles less than 12 degrees; subset 2 has 12 images with angles between 20 and 25 degrees; subset 3 contains 12 images with angles between 35 and 50 degrees; subset 4 has 14 images with angles between 60 and 77 degrees; and, finally, subset 5 contains 19 images with angles larger than 77 degrees. In other words, illumination variations become stronger by increasing the subset number.

\begin{figure}[t]
\centering
\vspace{0.00 in}
\includegraphics[width=0.91\linewidth]{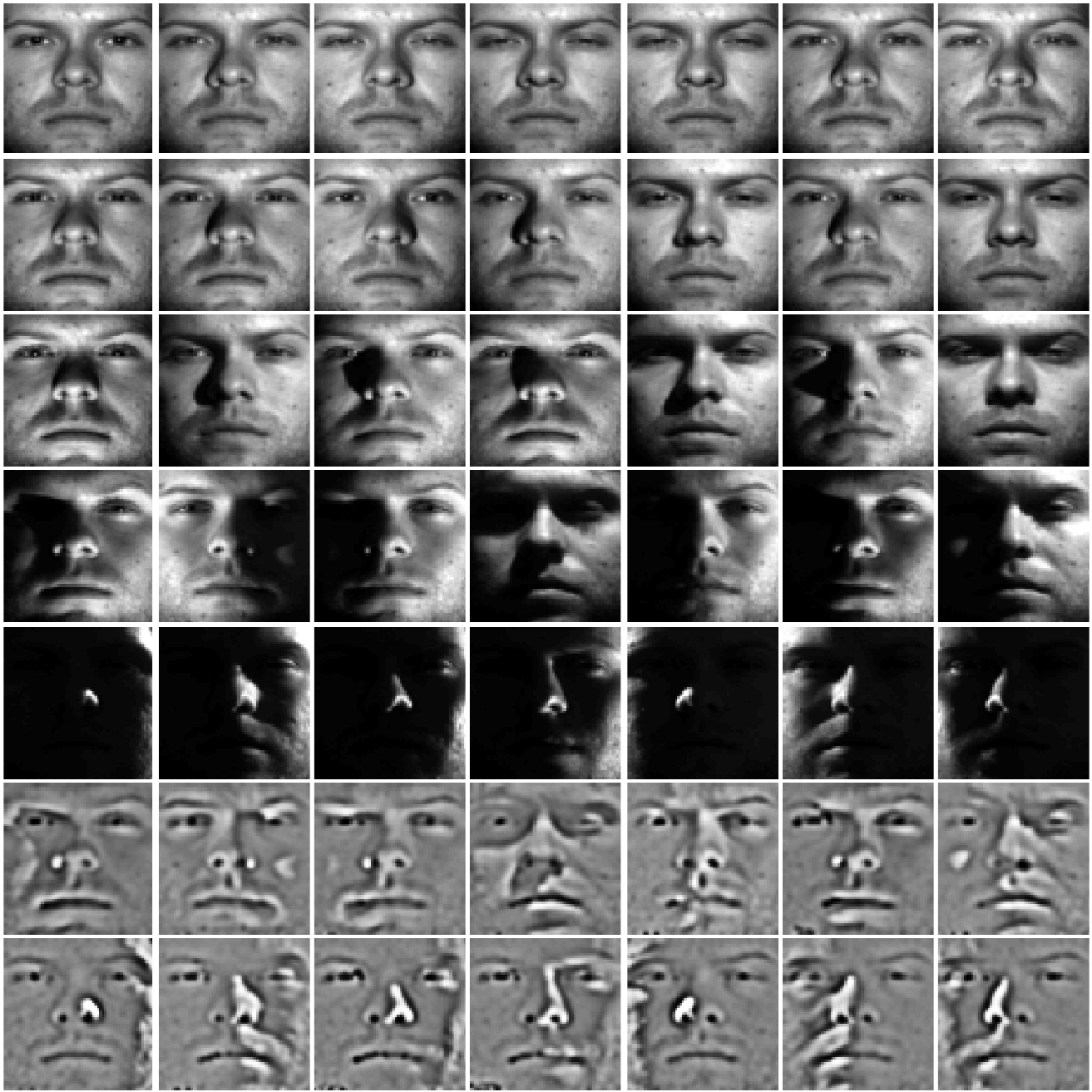}
\caption{Samples from the Extended Yale Dataset B with various illumination conditions. Rows 1 to 5 correspond to subsets 1 to 5, respectively, and the last two rows are the preprocessed samples of subsets 4 and 5.}
\label{fig3}
\end{figure}

To evaluate the effects of illumination variations using deep CNN-based features, only the frontal face images of this dataset under all illumination variations are selected. The first subset with almost perfect frontal illumination is used for training while subsets 2 to 5 are used for testing. All obtained images are later aligned, cropped into a square facial patch, and finally scaled to either $224 \times 224$ or $128 \times 128$ pixels. The VGG-Face mean image is then subtracted from each image. A few samples associated with one subject from the Extended Yale database B are shown in \Cref{fig3}. The results of the experiments on illumination variations using the Extended Yale dataset B subsets are reported in \Cref{table3}.

\begin{table}[b]
\centering
\vspace{-0.05 in}
\small
\caption{Classification results (\%) for the Extended Yale database B using deep representations against various illumination conditions}
\label{table3}
\renewcommand{\arraystretch}{1.2}
\begin{tabular}{cccc} \hline
\textbf{Testing Set} & \multicolumn{2}{c}{\textbf{VGG-Face}} & \textbf{Lightened CNN} \\
 & \textbf{FC6} & \textbf{FC7} & \\ \hline \hline
\text{Subset 2} & \textbf{100} & \textbf{100} & 82.43 (A) \\
\text{Subset 3} & 88.38 & \textbf{92.32} & 18.42 (B) \\
\text{Subset 4} & 46.62 & \textbf{52.44} & 8.46 (B) \\
\text{Subset 5} & 13.85 & \textbf{18.28} & 4.29 (B) \\
\text{Preprocessed Subset 4} & 71.80 & \textbf{75.56} & 26.32 (A) \\
\text{Preprocessed Subset 5} & 73.82 & \textbf{76.32} & 24.93 (A) \\ \hline
\end{tabular}
\end{table}

As it can be observed, deep face representations are robust against small illumination variations which exist in subsets 2 and 3. However, the performance degrades significantly when the illumination change strength increases. The main reason for this outcome can be attributed to the fact that the deep face models are mainly trained on celebrity pictures obtained from the web that are usually collected under relatively well-illuminated conditions. Therefore, they do not learn to handle strong illumination variations. One way to tackle this problem would be to employ preprocessing before applying the deep CNN models for feature extraction. To assess the contribution of image preprocessing, we preprocessed face images of subsets 4 and 5 by illumination normalization and contrast enhancement and ran the same experiments on these newly obtained subsets. The corresponding results are shown in the last two rows of \Cref{table3}. As it can be seen, preprocessing helps to improve the obtained accuracies. These results justify that, although deep CNNs provide a powerful representation for face recognition, they can still benefit from the preprocessing approaches. This is especially the case, when the variations available in the test set are not accounted for pre-training, making it essential to normalize these variations.

\subsection{Color FERET Database -- Pose Variations}
The color FERET database \cite{feret1998} contains 11,338 color images of size $512 \times 768$ pixels captured in a semi-controlled environment with 13 different poses from 994 subjects. To benchmark robustness of deep features against pose variations, we use the regular frontal image set (\textit{fa}) for training with one frontal image per subject. The network is then tested on six non-frontal poses, including two quarter left (\textit{ql}) and quarter right (\textit{qr}) poses with head tilts of about 22.5 degrees to left and right, two half left (\textit{hl}) and half right (\textit{hr}) poses with head tilts of around 67.5 degrees, and two profile left (\textit{pl}) and profile right (\textit{pl}) poses with head tilts of around 90 degrees.

\begin{figure}[t]
\centering
\vspace{0.00 in}
\includegraphics[width=0.91\linewidth]{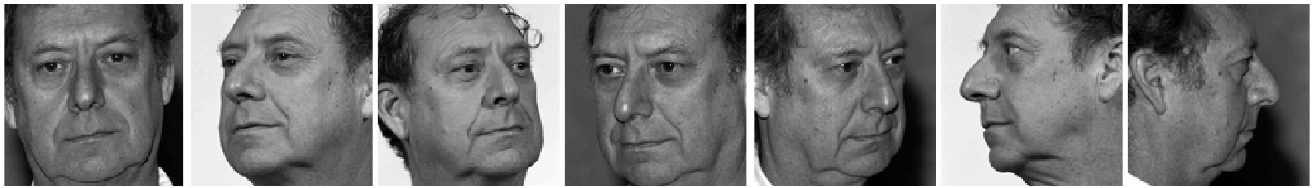}
\caption{Samples from the color FERET database with different pose conditions. The first image on the left is the frontal face picture (\textit{fa}) used for training and the rest (\textit{ql, qr, hl, hr, pl, pr}) are assigned for testing.}
\label{fig4}
\end{figure}

All the utilized images are cropped into a square facial patch and scaled to either $224 \times 224$ or $128 \times 128$ pixels. The VGG-Face mean image is then subtracted from each image. \Cref{fig4} shows samples associated with one subject from the FERET database. The obtained accuracies for pose variations on the same datasets are reported in \Cref{table4}.

\begin{table}[b]
\centering
\vspace{-0.03 in}
\small
\caption{Classification results (\%) for the FERET database using deep features against different pose conditions}
\label{table4}
\renewcommand{\arraystretch}{1.2}
\begin{tabular}{cccc} \hline
\textbf{Testing Set} & \multicolumn{2}{c}{\textbf{VGG-Face}} & \textbf{Lightened CNN} \\
 & \textbf{FC6} & \textbf{FC7} & \\ \hline \hline
\text{Quarter Left} & \textbf{97.63} & 96.71 & 25.76 (A) \\
\text{Quarter Right} & \textbf{98.42} & 98.16 & 26.02 (A) \\
\text{Half Left} & \textbf{88.32} & 87.85 & 6.08 (B) \\
\text{Half Right} & \textbf{91.74} & 87.85 & 5.98 (A) \\
\text{Profile Left} & 40.63 & \textbf{43.60} & 0.76 (B) \\
\text{Profile Right} & 43.95& \textbf{44.53} & 1.10 (B) \\ \hline
\end{tabular}
\end{table}

As the results indicate, the VGG-Face model is able to handle pose variations of up to 67.5 degrees. Nevertheless, the results can be further improved by employing pose normalization approaches, which have been already found useful for face recognition \cite{gao2009, asthana2011, hassner2015}. The performance drops significantly when the system is tested with profile images. Besides the fact that frontal-to-profile face matching is a challenging problem, the lack of enough profile images in the training datasets of deep CNN face models could be reason behind this performance degradation.

\subsection{The FRGC Database -- Misalignment}
The Face Recognition Grand Challenge (FRGC) database \cite{frgc2005} contains frontal face images photographed both in controlled and uncontrolled environments under two different lighting conditions with neutral or smiling facial expressions. The controlled subset of images was captured in a studio setting, while the uncontrolled photographs were taken either in hallways, atria, or outdoors.

To assess the robustness of deep CNN-based features against misalignment, the Fall 2003 and Spring 2004 collections are utilized and divided into controlled and uncontrolled subsets to obtain four new subsets, each containing photographs of 120 subjects with ten images per subject. The Fall 2003 subsets are used for gallery, while those from Spring 2004 are employed as probe images. In other words, gallery images are from the controlled (uncontrolled) subset of Fall 2003 and probe images are from the controlled (uncontrolled) subset of Spring 2004. We named the experiments run under controlled conditions FRGC1 and those conducted under uncontrolled conditions, FRGC4.

\begin{figure}[t]
\centering
\vspace{0.01 in}
\includegraphics[width=0.91\linewidth]{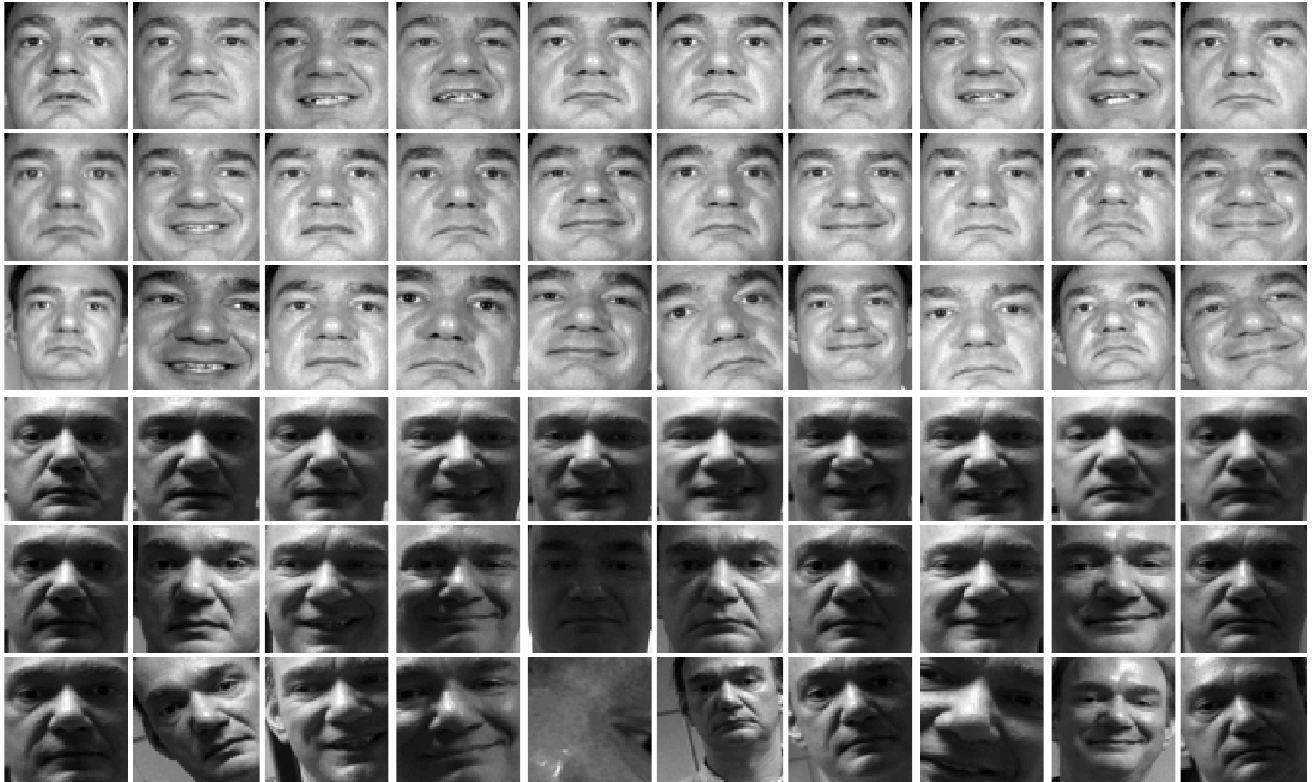}
\caption{Samples from the FRGC database aligned with different registration errors. The first three rows are acquired from FRGC1 and show the training samples (row1) and testing samples aligned with zero (row2) and 10\% (row3) registration errors, respectively. The second three rows are associated with FRGC4 and depict the training samples (row4) and testing samples aligned with zero (row5) and 20\% (row6) registration errors, respectively.}
\label{fig5}
\end{figure}

\begin{figure}[t]
\centering
\vspace{0.00 in}
\includegraphics[width=0.97\linewidth]{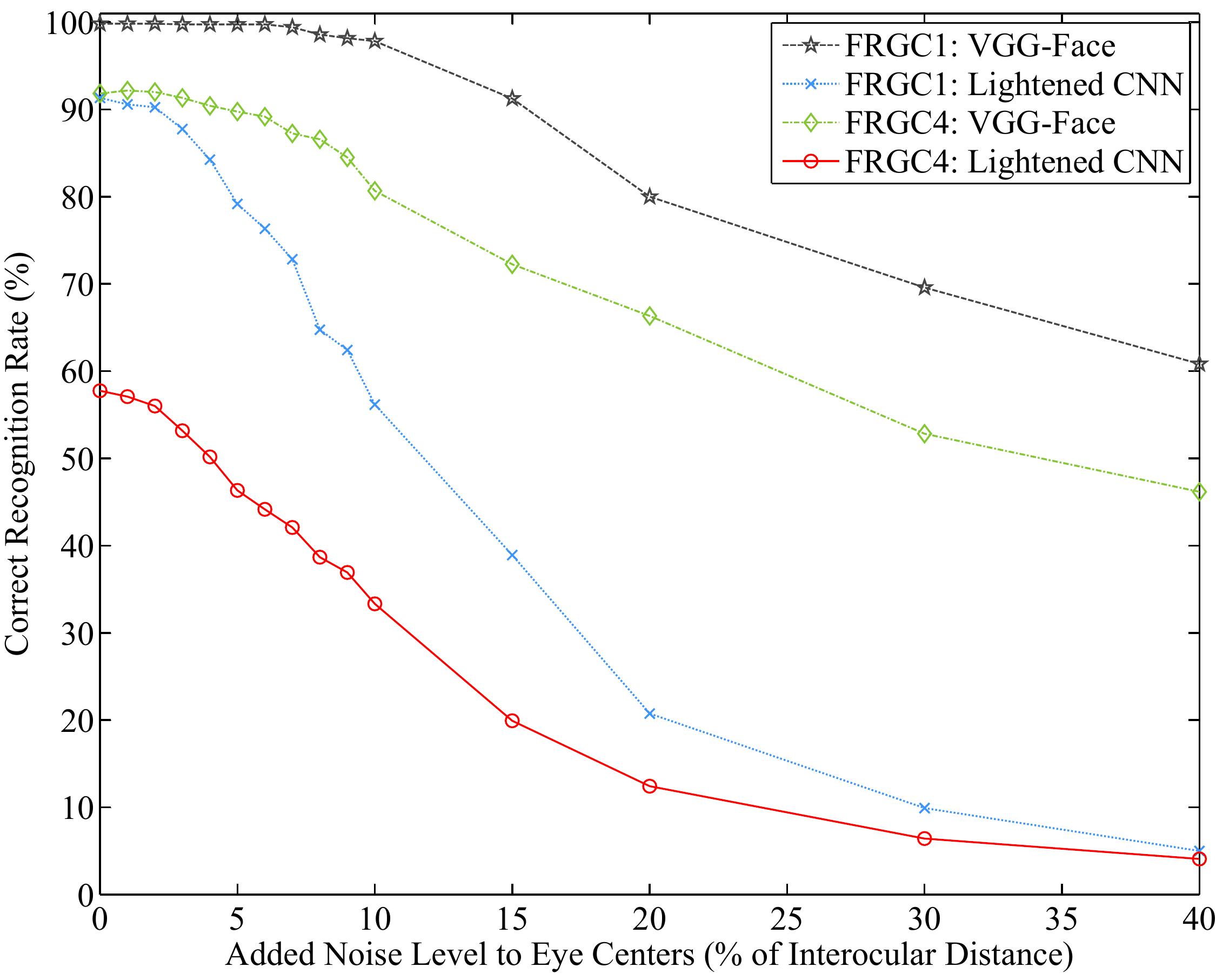}
\caption{Classification results for the FRGC dataset using deep facial representations against different facial feature localization error levels}
\label{fig6}
\end{figure}

Similar to previous tasks, the gallery images are aligned with respect to manually annotated eye center coordinates, cropped into a square facial patch, and scaled to either $224 \times 224$ or $128 \times 128$ pixels. The VGG-Face mean image is also subtracted from each image. To imitate misalignment due to erroneous facial feature localization, while aligning probe images, random noise up to 40\% of the distance between the eyes is added to the manually annotated eye center positions. \Cref{fig5} shows sample face images associated with one subject from the different subsets of the FRGC database. The classification results of the utilized deep models with respect to varying degrees of facial feature localization errors are shown in \Cref{fig6}. Note that the VGG-Face features for this task are all obtained from FC6.

Analysis of the results displayed in \Cref{fig6} shows that deep CNN-based face representation is robust against misalignment, i.e. it can tolerate up to 10\% of interocular distance error from the facial feature localization systems. This is a very important property since traditional appearance-based face recognition algorithms have been known to be sensitive to misalignment.

\subsection{Facial Bounding Box Extension}
As our last experiment, we evaluate deep facial representations against alignment with a larger facial bounding box. For this purpose, each image of the utilized datasets is aligned and cropped into an extended square facial patch to include all parts of the head, i.e. ears, hair, and the chain. These images are then scaled to either $224 \times 224$ or $128 \times 128$ pixels and the VGG-Face mean image is subtracted from each image. \Cref{table5} shows the results of alignment with larger bounding boxes on different face datasets.

\begin{table}[t]
\centering
\vspace{0.00 in}
\small
\caption{Classification results (\%) using deep features for different face datasets aligned with a larger bounding box}
\label{table5}
\renewcommand{\arraystretch}{1.2}
\begin{tabular}{ccc} \hline
\textbf{Training Set} & \textbf{Testing Set} & \textbf{VGG-Face (FC6)} \\ \hline \hline
\text{AR Neutral Set 1} & \text{AR Sunglasses Set 1} & 44.55 \\
\text{AR Neutral Set 1} & \text{AR Scarf Set 1} & 93.64 \\
\text{AR Neutral Set 1} & \text{AR Sunglasses Set 2} & 39.09 \\
\text{AR Neutral Set 1} & \text{AR Scarf Set 2} & 91.82 \\
\text{CMU PIE Train} & \text{CMU PIE Test} & 97.72 \\
\text{Ext. Yale Set 1} & \text{Ext. Yale Set 2} & 100 \\
\text{Ext. Yale Set 1} & \text{Ext. Yale Set 3} & 94.52 \\
\text{Ext. Yale Set 1} & \text{Ext. Yale Set 4} & 56.58 \\
\text{Ext. Yale Set 1} & \text{Ext. Yale Set 5} & 27.56 \\ \hline
\end{tabular}
\end{table}

Comparing the obtained results in \Cref{table5} with those of \Cref{table1,table2,table3} shows that using deep features extracted from the whole head remarkably improves the performance. One possible explanation for this observation is that the VGG-Face model is trained on images that contained all the head rather than merely the face image; therefore, extending the facial bounding box increases classification accuracy by including useful features extracted from the full head.

\section{Summary and Discussion} \label{conc_sec}
In this paper, we presented a comprehensive evaluation of deep learning based representation for face recognition under various conditions including pose, illumination, occlusion, and misalignment. Two successful deep CNN models, namely VGG-Face \cite{vggface2015} and Lightened CNN \cite{lightened2015}, pre-trained on very large face datasets, were employed to extract facial image representations. Five well-known face datasets were utilized for these experiments, namely the AR face database \cite{ar1998} to analyze the effects of occlusion, CMU PIE \cite{pie2002} and Extended Yale dataset B \cite{extyale2001} for analysis of illumination variations, Color FERET database \cite{feret1998} to assess the impacts of pose variations, and the FRGC database \cite{frgc2005} to evaluate the effects of misalignment.

It has been shown that deep learning based representations provide promising results. However, the achieved performance levels are not as high as those from the state-of-the-art methods reported on these databases in the literature. The performance gap is significant for the cases in which the tested conditions are scarce in the training datasets of CNN models. We propose that using preprocessing methods for pose and illumination normalization along with pre-trained deep learning models or accounting for these variations during training substantially resolve this weakness. Besides these important observations, this study has revealed that an advantage of deep learning based face representations is their robustness to misalignment since they can tolerate misalignment due to facial feature localization errors of up to 10\% of the interocular distance.

The VGG-Face model has shown a better transferability compared to the Lightened CNN model. This could be attributed to its more sophisticated architecture that results in a more abstract representation. On the other hand, the Lightened CNN model is, as its name implies, a faster approach that uses an uncommon activation function (MFM) instead of ReLU. Also, the VGG-Face features obtained from the FC6 layer show better robustness against pose variations, while those obtained from the FC7 layer have better robustness to illumination variations.

Overall, although a significant progress has been achieved during the recent years with the deep learning based approaches, face recognition under mismatched conditions--especially when a limited amount of data is available for the task at hand--still remains a challenging problem.

\subsection*{Acknowledgements}
This work was supported by TUBITAK project no. 113E067 and by a Marie Curie FP7 Integration Grant within the 7th EU Framework Programme.

{\small
\bibliographystyle{ieee}
\bibliography{refs}
}

\end{document}